\documentclass[twoside]{article}
\usepackage{ecj,palatino,epsfig,latexsym,natbib}

\usepackage{hyperref}
\usepackage{graphicx}
\usepackage{xargs}
\usepackage[pdftex,dvipsnames]{xcolor}
\usepackage{multirow}
\usepackage{booktabs}
\usepackage{algorithm}
\usepackage{algpseudocode}
\usepackage{varwidth}
\usepackage{scrextend}
\usepackage{amsmath}
\usepackage{amssymb}

\newcommand{\ie}{i.e.}
\newcommand{\eg}{e.g.}
\newcommand{\vs}{vs.}

\newcommand{\citeg}[1]{\citep[see, \eg{},][]{#1}}

\begin{document}

\ecjHeader{x}{x}{xxx-xxx}{201X}{Automated Algorithm Selection: Survey and Perspectives}{P. Kerschke, H. H. Hoos, F. Neumann, H. Trautmann}
\title{Automated Algorithm Selection: \\Survey and Perspectives}

\author{\name{\bf Pascal Kerschke} \hfill \addr{kerschke@uni-muenster.de}\\ 
        \addr{Information Systems and Statistics, University of M{\"u}nster, 48149 M{\"u}nster, Germany}
\AND
	\name{\bf Holger H. Hoos} \hfill \addr{hh@liacs.nl}\\ 
        \addr{Leiden Institute of Advanced Computer Science, Leiden University, 
        2333 CA Leiden, The Netherlands}
\AND
	\name{\bf Frank Neumann} \hfill \addr{frank.neumann@adelaide.edu.au }\\ 
        \addr{Optimisation and Logistics, The University of Adelaide, Adelaide, SA 5005, Australia}
\AND
	\name{\bf Heike Trautmann} \hfill \addr{trautmann@uni-muenster.de}\\ 
        \addr{Information Systems and Statistics, University of M{\"u}nster, 48149 M{\"u}nster, Germany}
}

\maketitle

\begin{abstract}

It has long been observed that for practically any computational problem that has been intensely studied, different instances are best solved using different algorithms. 
This is particularly pronounced for computationally hard problems, where in most cases, no single algorithm defines the state of the art; instead, there is a set of algorithms with complementary strengths.
This performance complementarity can be exploited in various ways, one of which is based on the idea of selecting, from a set of given algorithms, for each problem instance to be solved the one expected to perform best.
The task of automatically selecting an algorithm from a given set is known as the \emph{per-instance algorithm selection problem} and has been intensely studied over the past 15 years, leading to major improvements in the state of the art in solving a growing number of discrete combinatorial problems, including propositional satisfiability and AI planning.
Per-instance algorithm selection also shows much promise for boosting performance in solving continuous and mixed discrete/continuous optimisation problems.

This survey provides an overview of research in automated algorithm selection, ranging from early and seminal works to recent and promising application areas. 
Different from earlier work, it covers applications to discrete and continuous problems, and discusses algorithm selection in context with conceptually related approaches, such as algorithm configuration, scheduling or portfolio selection. 
Since informative and cheaply computable problem instance features provide the basis for effective per-instance algorithm selection systems, we also provide an overview of such features for discrete and continuous problems. 
Finally, we provide perspectives on future work in the area and discuss a number of open research challenges.

\end{abstract}


\section{Introduction}
\label{sec:intro}

It has long been observed that for well-studied computational problems for which several high-performance algorithms are available, there is typically no single algorithm that dominates
all others on all problem instances.
Instead, different algorithms perform best on different types of problem instances
-- a phenomenon also known as \emph{performance complementarity}, which is often incorrectly
attributed to an interesting theoretical result known as the \textit{no-free-lunch (NFL) theorem} \citep{wolpert1995,wolpert1997}. In the context of search problems, the NFL theorem strictly only applies if arbitrary search landscapes are considered, while the instances of basically any search problem of interest have compact descriptions and therefore cannot give rise to arbitrary search landscapes \citep{culberson1998}.
Performance complementarity has been observed for practically all NP-hard decision and optimisation problems;
these include propositional satisfiability, constraint satisfaction, a wide range of planning and scheduling problems, mixed integer programming and the travelling salesperson problem, as well as a broad range of continuous optimisation, machine learning and important polynomial-time-solvable problems (\eg{}, sorting and shortest path finding).
For these and many other problems, theoretical results stating which algorithmic strategies work best are restricted to very limited classes of problem instances, so that it is generally unknown \emph{a priori} which of several algorithms should be used to solve a given instance.

This gives rise to the increasingly prominent \emph{per-instance algorithm selection problem}: 
given a computational problem, a set of algorithms for solving this problem, and a specific instance that needs to be solved, determine which of the algorithms can be expected to perform best on that instance.
This problem has already been considered in the seminal work by \citet{rice1976}, but it took several decades before practical per-instance algorithm selection methods became available \citeg{cook1997,leytonbrown2003,xu2008}.
Since then, the problem and algorithms for solving it have steadily gained prominence, and by now have given rise to a large body of literature.
Indeed, per-instance algorithm selection techniques have produced substantial improvements in the state of the art in solving a large range of prominent computational problems, including propositional satisfiability (SAT) and the travelling salesperson problems (TSP) \citep{xu2008,xu2012,kerschke2017_tsp}.

It is important to note that there are several concepts that are quite closely related to that of per-instance algorithm selection, notably, per-set algorithm selection, algorithm configuration, algorithm schedules and parallel algorithm portfolios, which are all discussed in further detail in Section~\ref{sec:related}. 
Unfortunately, there is some potential for confusion, especially between per-instance algorithm selection and parallel algorithm portfolios, since in the literature, the term \emph{portfolio} is sometimes used to refer to algorithm selectors. 
Furthermore, some of the most prominent and successful algorithm selection approaches from the literature, such as \textsc{SATzilla} \citep{xu2008} and \textsc{Auto\-Folio} \citep{lindauer2015}, implement combinations of algorithm scheduling and per-instance selection.
While we will briefly discuss these more complex systems, along with approaches that select more than one algorithm to be run on a given problem instance, the focus of this survey is on pure per-instance algorithm selection, as outlined above and defined formally in Section~\ref{sec:related}.

We note that per-instance algorithm selection can be applied to optimisation problems, where the goal is to find an optimal (or best possible) solution according to a given objective function, as well as to decision problems, where one wants to determine, as quickly as possible, whether a solution satisfying certain conditions exists.
Furthermore, it is useful to distinguish between continuous problems, where the components of a possible solution are real numbers (possibly constrained to a given interval), and discrete problems, where candidate solutions are discrete objects, such as graphs, permutations or vectors of integers.

Several surveys on algorithm selection have been published over the last decade. 
In the first extensive survey in this area, \citet{smithmiles2009} summarised developments in the meta-learning, artificial intelligence and operations research communities. Adopting a cross-disciplinary perspective, she combined contributions from these areas under the umbrella of a ``\mbox{(meta-)}learning'' framework, which permitted her to identify parallel and closely related developments within these rather well-separated communities.
However, this survey was published a decade ago and therefore does not cover recent developments and improvements to the state of the art in this fast-moving research area.

A more recent overview on algorithm selection was published by \cite{kotthoff2014}. 
His survey presents an extensive, valuable guide to the automated algorithm selection literature up to 2014 and provides answers to several important questions, such as (i) what are the differences between static and dynamic portfolios, (ii) what should be selected (single solver, schedule, different candidate portfolios), (iii) what are the differences between online and offline selection, (iv) how should the costs for using algorithm portfolios be considered, (v) which prediction type (classification, regression, etc.) is most promising when training an algorithm selector, and (vi) what are differences between static and dynamic, as well as low-level and high-level features. 
Unfortunately, Kotthoff's survey is restricted to algorithm selection for discrete problems and does not cover in any detail problem instance features, which provide the basis for per-instance algorithm selection.

Those two limitations where -- at least partially -- addressed by \citet{munoz2013}. 
Although the title (``The Algorithm Selection Problem on the Continuous Optimization Domain'') appears to suggest otherwise, their survey mostly addresses the paucity of work on algorithm selection for continuous optimisation problems and the challenges arising in this context. 
Rather than providing an overview of algorithm selection approaches in this area, \citet{munoz2013} summarise promising results on discrete problems and hint at the possibility of achieving similar results in continuous optimisation. 
In their follow-up survey, \cite{munoz2015_as} provide further insights into the existing ingredients for algorithm selection in the domain of continuous optimisation: benchmarks, algorithms, performance metrics, and problem characteristics obtained by exploratory landscape analysis. Still, they do not cover any work describing automated algorithm selection in this domain.

Our goal here is to not only update, but also to complement and extend these previous surveys.
Firstly, we cover work on algorithm selection for discrete \textit{and} continuous problems; 
as a result, we can compare the difficulties, challenges and solutions found in those domains.
Secondly, one of the most important ingredients for successful algorithm selection approaches are informative (problem-specific) features. 
We therefore provide an overview of several promising feature sets and discuss characteristics that have been demonstrated to provide a strong basis for algorithm selection.
Thirdly, we discuss several problems closely related to (and sometimes confused with) algorithm selection, such as automated algorithm configuration, algorithm schedules and parallel portfolios, pointing out differences, similarities and synergies.
Of course, in light of the considerable and fast-growing body of literature on and related to algorithm selection, we cannot provide comprehensive coverage; instead, we selected contributions based on their impact, promise and conceptual contributions to the area.

The remainder of this survey article is structured as follows.
In Section~\ref{sec:related}, we formally define the per-instance algorithm selection problem and situate it in the context of related problems, such as automated algorithm configuration. Next, Section~\ref{sec:features} provides an overview of instance features for discrete and continuous optimisation problems that provide the basis for automated algorithm selection. 
Successful applications of algorithm selection in discrete and continuous optimisation are discussed in Sections~\ref{sec:discrete} and \ref{sec:continuous}, respectively.
Finally, Section~\ref{sec:perspectives} provides additional perspectives on algorithm configuration and outlines several open challenges.

\begin{figure}[!t]
\centering
\includegraphics[width=\textwidth, trim = 8mm 26mm 15mm 16mm, clip]{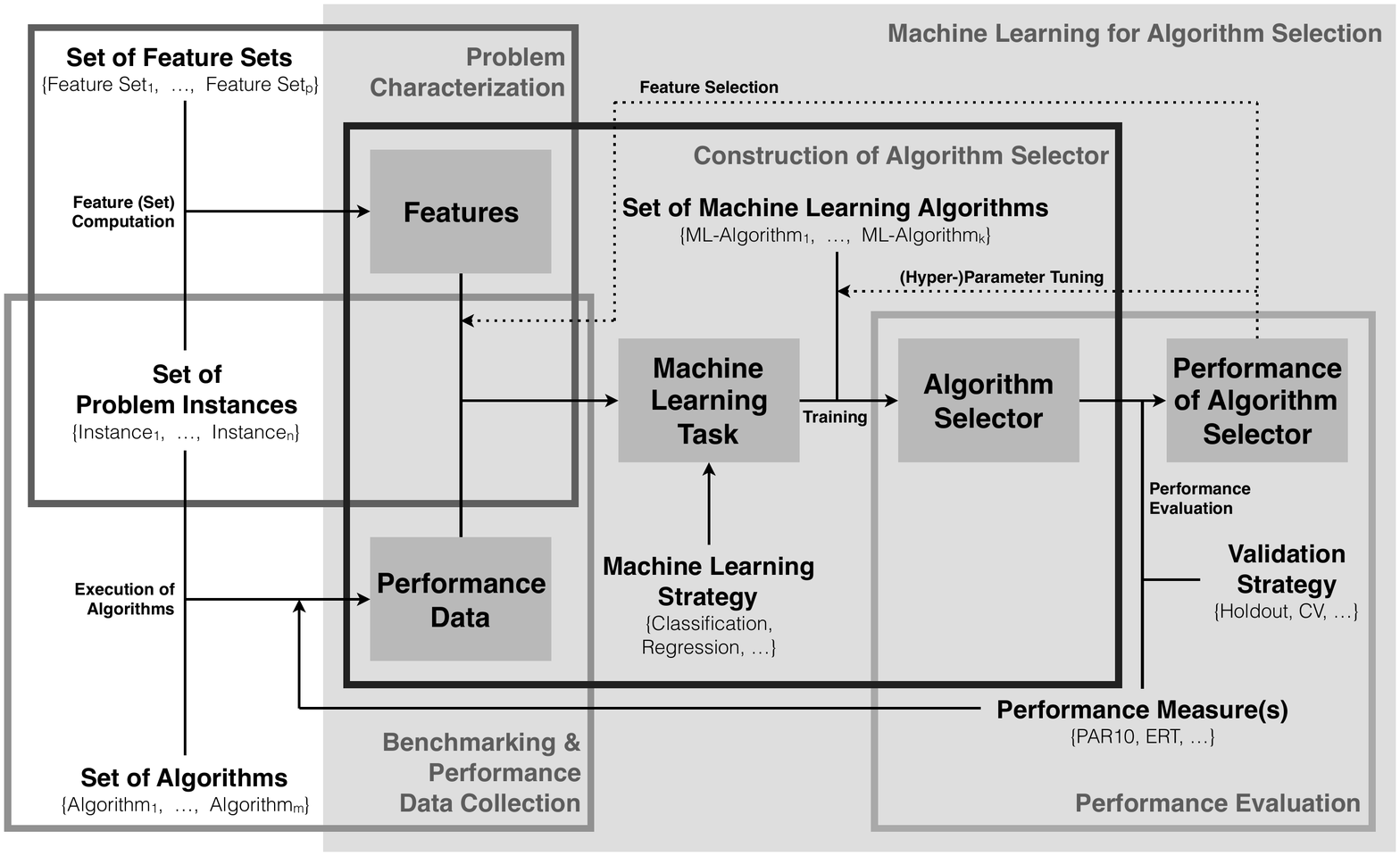}
\caption{Schematic overview of interplay between problem instance features (top left), algorithm performance data (bottom left), selector construction (center) and the assessment of selector performance (bottom right). }
\label{fig:scheme_overview}
\end{figure}

\begin{figure}[!t]
\centering
\includegraphics[width=\textwidth, trim = 22mm 11mm 13mm 46mm, clip]{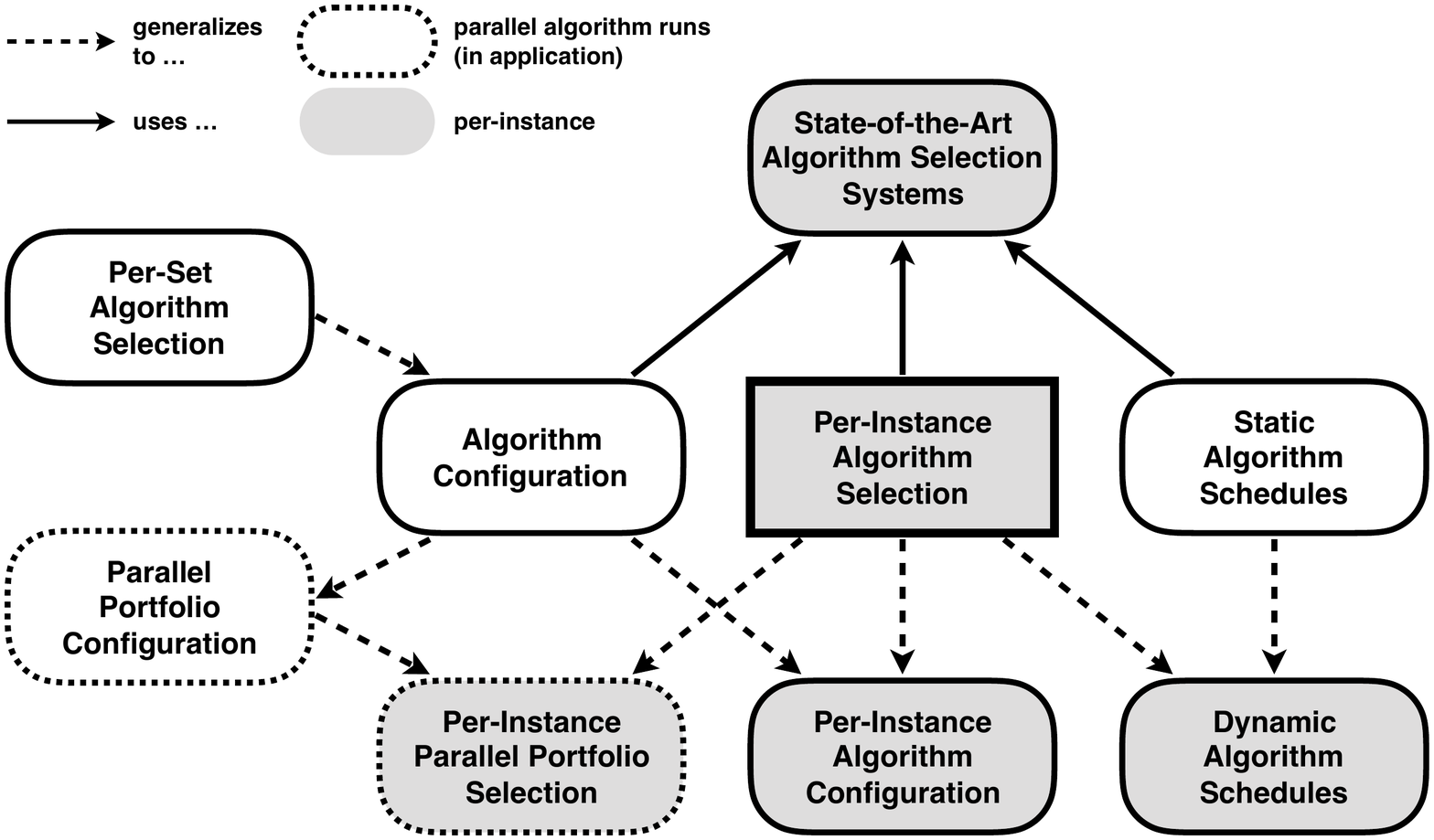}
\caption{Connections between per-instance algorithm selection and related problems.}
\label{fig:scheme_related}
\end{figure}

\section{Algorithm Selection and Related Problems}
\label{sec:related}

We consider the selection of algorithms for a given decision or optimisation problem $P$.
Specifically, the per-instance algorithm selection problem can be formulated as follows \citep[see also][]{rice1976}:
Given a set $I$ of instances of a problem $P$, a set $\mathbf{A} = \{A_1, \ldots A_n\}$ of algorithms for $P$
and a metric $m: \mathbf{A} \times I \rightarrow \mathbb{R}$ that measures the performance of any algorithm $A_j \in \mathbf{A}$ on instance set $I$,
construct a selector $S$ that maps any problem instance $i \in I$ to an algorithm $S(i) \in \mathbf{A}$ such that the overall performance of $S$ on $I$ is optimal according to metric $m$.

Of course, in general, we cannot hope to efficiently find perfect solutions to the per-instance algorithm selection problem, and instead, we strive to find selectors whose performance is as close as possible to that of a perfect selector on instance set $I$. This is typically achieved by making use of informative and cheaply computable features 
$\mathbf{f}(i) = (f_1(i), \ldots, f_k(i))$ of the given problem instance $i$. 
A general overview on the interplay of instance features, algorithm performance data, and algorithm selection is shown in Figure~\ref{fig:scheme_overview}.
The features are of key importance, and we will discuss them in more detail in the following section of this article.  

We note that the performance of a (hypothetical) perfect per-instance algorithm selector, often also referred to as an \emph{oracle selector}  or \emph{virtual best solver (VBS)}, provides a lower bound on the performance of any realistically achievable algorithm selector (where we assume, w.l.o.g., that the given performance measure is to be minimised), and is often used in the context of assessing selector performance.

Another useful concept is that of the \emph{single best solver (SBS)}, which is the algorithm $A'$ with the best performance among all the $A_j \in \mathbf{A}$. The SBS is the solution to the closely related \emph{per-set algorithm selection problem}, and its performance provides a natural upper bound on the performance of any reasonable per-instance algorithm selector. Furthermore, the difference or ratio between the performance of the SBS and VBS, also known as the \emph{VBS-SBS gap}, gives an indication of the performance gains that can be realised, in the best case, by per-instance algorithm selection, and the fraction of the VBS-SBS gap closed by any per-instance algorithm selector $S$ provides a measure of its performance \citeg{lindauer2017}.
State-of-the-art per-instance algorithm selectors for combinatorial problems have demonstrated to close between 25\% and 96\% of the VBS-SBS gap \citeg{lindauer2015}.
It is important to note that the VBS-SBS gap is large when the given set $\mathbf{A}$ of algorithms shows high performance complementarity on instance set $I$, \ie{}, when different $A_j \in \mathbf{A}$ perform best on different $i \in I$, and those algorithms that are best on some instances perform quite poorly on others.
Generally, per-instance algorithm selection can be expected to achieve large performance gains over the single best algorithm if there is high performance complementarity within $\mathbf{A}$ and there is a set of sufficiently cheaply computable and informative instance features that can be leveraged in learning a good mapping from instances to algorithms.

It is very important to distinguish between per-set algorithm selection and per-instance algorithm selection. The former does not require any instance features and is typically done by exhaustive evaluation of all given algorithms on a set of problem instances deemed to be representative for those to be solved later. The connection between per-instance algorithm selection and related problems is shown in Figure~\ref{fig:scheme_related}.
In many ways, algorithm competitions, such as the international SAT and planning competitions
\citeg{jaervisalo2012,vallati2015}, 
can be seen as identifying solutions to per-set algorithm selection problems for broad sets of interesting instances,
and competition winners are often seen as the single best algorithm for the respective problem.
Sometimes, to reduce the computational cost for per-set algorithm selection, racing methods are used.
These run candidate algorithms on an increasing number of instances, eliminating those from consideration whose performance is significantly below that of others, based on a statistical test \citeg{maron1994,birattari2002}.

Per-set algorithm configuration can be seen as a special case of \emph{algorithm configuration}, a practically very important problem that can be described as follows: Given an algorithm $A$ whose performance (but not semantics) is affected by the settings of parameters $\mathbf{p} = (p_1, \ldots, p_k)$, a set $C$ of possible values for $\mathbf{p}$ (called \emph{configurations} of $A$), a set of problem instances $I$ and a performance metric $m$, find a configuration $c^\ast \in C$ of $A$ that achieves optimal performance on $I$ according to $m$.
Note that the set $C$ of configurations can be seen as corresponding to a set of algorithms, of which we wish to select the one that performs best. The key difference to algorithm selection is that this set can be very large, since it arises from combinatorial combinations of values of the individual parameters $p_l$, which, in some cases, can take continuous values, leading to (potentially) uncountably infinite sets of algorithm configurations over which we have to optimise. Realistic algorithm configuration scenarios typically involve tens to hundreds of parameters (see, \eg{}, \citealp{hutter2009, hutter2011,lopez2016,ansotegui2015, thornton2013, kotthoff2017}).
Therefore, per-set algorithm selection techniques are typically not directly applicable to algorithm configuration, although racing techniques can be extended to work well in this case \citeg{lopez2016,caceres2017}.
Per-set algorithm configuration is closely related to hyperparameter optimisation in machine learning; the main difference is that in algorithm configuration, performance is to be optimised on a possibly diverse set of problem instances, which often requires trading off performance on some instance against that achieved on others. In a sense, the typical hyperparameter optimisation problem encountered in machine learning is analogous to configuring a parameterised algorithm for performance on a single problem instance.

Since typical procedures used for building per-instance algorithm selectors have design choices that can be exposed as parameters, algorithm configuration techniques can be applied to optimise their performance on specific (sets of) selection scenarios. This has been done, with considerable success, in the recent \textsc{AutoFolio} selection system by \citet{lindauer2015}, which we will discuss in further detail in Section~\ref{sec:discrete}.

The per-instance variant of the algorithm configuration problem, which can be seen as a generalisation of per-instance algorithm selection, largely remains an open challenge \citeg{hutter2006,belkhir2016,belkhir2017}, and we briefly discuss it further in Section~\ref{sec:perspectives}.

Performance complementarity within a set of algorithms can be leveraged in ways that differ from per-instance algorithm configuration. 
One prominent approach is that of a \emph{parallel algorithm portfolio}, where each algorithm from a given set 
$\mathbf{A}$ is run in parallel on a given problem instance $i$ (see, \eg{}, \citealp{huberman1997,gomes2001,fukunaga2000}).
When applied to a decision problem, all runs are terminated as soon as one of the component algorithms has solved the given instance $i$; for optimisation problems, the best solution achieved by any of the component algorithms at any given time is returned as the solution of the entire portfolio.
Parallel algorithm portfolios are conceptually similar to ensemble methods in machine learning \citeg{dietterich2000,rokach2010}.
The key difference is that ensemble methods aggregate the results from the various component algorithms, \eg{}, by weighted or unweighted averaging.

When run on parallel hardware, algorithm portfolios typically achieve performance very close to that of the VBS in terms of wall-clock time, at the price of parallelism of degree equal to the number $n$ of algorithms in $\mathbf{A}$.
Of course, parallel portfolios can be run at lower actual degrees of parallelism, and even fully sequentially, using task-switching, as provided, \eg{}, by the operating system; in that case, the wall-clock time is typically close to that of $n$ times the performance of the VBS in terms of the time required to solve a given instance of a decision problem, or to achieve a certain solution quality in case of an optimisation problem. Most of this overhead, which for large sets of algorithms can be very substantial, can be avoided by using per-instance algorithm selectors instead of parallel portfolios.
Nevertheless, especially in the area of evolutionary computation, the concept of parallel algorithm portfolios has given rise to a growing body of research, in which the basic concept is often combined with additional techniques to achieve improved performance \citeg{tang2014,yuen2015}.

Although the term \emph{algorithm portfolio} is sometimes used in the literature to refer to per-instance algorithm selectors and other techniques that leverage performance complementarity within a set of algorithms, we discourage this broad use as it easily leads to confusion between conceptually very different approaches.
This potential confusion is easily avoided by restricting the term \emph{algorithm portfolio} to parallel algorithm portfolios, consistent with the seminal work by \citet{huberman1997}.
Recently, the concepts of per-instance algorithm selection and parallel portfolios have been combined, by selecting, on a per-instance basis, several algorithms to be run in parallel \citep{lindauer2015_AS_to_PPS}. 
While this would not make sense in the context of a perfect algorithm selector, it can limit the impact of the poor selection decisions sometimes made in practice.

\emph{Algorithm schedules} provide another way of exploiting performance complementarity (see, \eg{}, \citealp{lindauer2014,lindauer2016}). The key idea is to run a sequence of algorithms from a given set $\mathbf{A}$, one after the other, each for a given (maximum) time. Those cut-off times can differ between the stages of the schedule, and some algorithms may not be run at all.
Static algorithm schedules, \ie{}, schedules that have been determined in a per-set fashion and are applied uniformly to any given problem instance $i$, can be quite effective and are typically much easier to implement than per-instance algorithm selectors \citeg{roussel2012}.
They are also used in state-of-the-art algorithm selection systems during a so-called pre-solving phase, in order to solve easy problem instances quickly and without the need for computing the instance features required for per-instance selection \citeg{xu2012,lindauer2015}.
Unless stated explicitly otherwise, we will in the following, when discussing per-instance algorithm selectors, always refer to the pure per-instance algorithm selection problem, as defined above, without pre-solving schedules and other extensions found in cutting-edge algorithm selection systems.

Finally, the problem of predicting the performance of an algorithm $A$ on a given problem instance $i$ is closely related to per-instance algorithm selection. 
If computationally cheap and accurate performance prediction were possible, evidently, we could use performance predictors for our given algorithms $A_1, \ldots, A_n$ and simply select the one predicted to perform best on $i$.
In practice, sufficient accuracy can be achieved for many problems using state-of-the-art regression techniques from machine learning, at moderate computational cost \citep{hutter2014}, and performance predictors form the basis for one of the main approaches to per-instance algorithm selection.
At the same time, other approaches, such as cost-based classification, exist and also find use in state-of-the-art algorithm selection systems, as explained in the following sections.

\section{Features for Discrete and Continuous Problems}
\label{sec:features}

Linking algorithm performance on an instance $i$ to instance characteristics forms a central part of automated algorithm selection and several related problems. For this purpose, automatically computable features $\mathbf{f}(i) = (f_1(i),\ldots,f_k(i))$ are required, ideally with the following properties: Firstly, features should be \emph{informative}, in that they allow for a sufficient distinction between different instances; they should also be \emph{interpretable}, so that feature values enable an expert to gain maximum insight into instance properties.
Furthermore, features should be \emph{cheaply computable}, so that the advantages gained by selecting an algorithm based on them is not outweighed by the cost of feature computation.
Features should also be \emph{generally applicable}, \ie{}, they should be effectively and efficiently computable for a broad range of problem instances, rather than being restricted, \eg{}, to small instance sizes.
Finally, the features $f_j$ should be \emph{complementary}, in that redundant sets of features are not only computationally wasteful, but can also cause problems when used by certain machine learning algorithms as a basis for algorithm selection and related problems.

In the following, we provide an overview of commonly used instance features for several prominent discrete and continuous problems -- not only to illustrate what kind of features are useful in the context of per-instance algorithm selection, but also to draw attention to an important and somewhat underrated research topic of significant importance to tasks beyond algorithm selection.
In particular, informative instance features can provide important insights into strengths and weaknesses of a given algorithm, and hence play a crucial role in devising improvements.
We generally distinguish between problem-specific features that are closely based on particular aspects of the problem to be solved, such as the number of clauses in instances of propositional satisfiability problems, and generic features that are more broadly applicable, such as high-level statistics over information gleaned from short 'probing' runs of a solver for the given problem.

%%%

\subsection{Discrete problems}
\label{sec:feats_discrete}

To give concrete examples, and in light of the importance of problem-specific features, we will focus on three of the most prominent and well-studied discrete combinatorial problems: propositional satisfiability (and related problems), AI planning and  the travelling salesperson problem (TSP).

\paragraph{Propositional satisfiability and related problems.}

The \textit{propositional satisfiability problem (SAT)} is to determine whether for a given formula $F$ in propositional logic, containing Boolean variables $X_1, \ldots, X_N \in \{true, false\}$, there exists an assignment of logical values to the variables such that $F$ evaluates to $true$; such a variable assignment is said to satisfy $F$.
Typically, the problem is restricted to formulae $F$ in \textit{conjunctive normal form (CNF)}, \ie{}, $F$ consists of conjunctions ($\wedge$) of so-called clauses, which are disjunctions ($\vee$) of Boolean variables $X_j$ and their negations $\neg X_j$.
A CNF-formula $F$ evaluates to $true$, if each of its clauses is satisfied simultaneously.
SAT is one of the most prominent and intensely studied combinatorial decision problems and has important applications in hard- and software verification \citeg{biere2009}.
Given the ties to other combinatorial problems, improvements in SAT often also impact widely studied related problems, such as \textit{maximum satisfiability (MaxSAT) problem}, in which the objective is to find a variable assignment that maximises the number of satisfied CNF clauses.

The first large collection of features for SAT (and thus also MaxSAT) instances was provided by \cite{nudelman2004}. Despite the rather simple structure of SAT instances, the authors devised nine different feature sets and a total of 91 features, which characterise a given CNF formula from a multitude of perspectives. 
Eleven \textit{problem size features} describe SAT instances based on summary statistics of their numbers of clauses and variables. 
A set of \textit{variable-clause graph (VCG) features} comprises ten node degree statistics based on a bipartite graph over the variables and clauses of a given instance. 
Interactions between the variables are captured by four \textit{variable graph (VG) features};
these are the minimum, maximum, mean and coefficient of variation of the node degrees for a graph of variables, 
in which edges connect pairs of variables that jointly occur in at least one clause. 
Similarly, the set of \textit{clause graph (CG) features} contains seven node degree statistics of a graph whose edges connect clauses that have at least one variable in common, as well as
three features based on weighted clustering coefficients for the clause graph. 
Thirteen \textit{balance features} capture the balance between negated and unnegated variables per clause, their overall occurrences across all clauses, as well as fractions of unary, binary and ternary clauses, whereas six further features quantify the degree to which the given $F$ resembles a Horn formula (a restricted type of CNF formula, for which SAT can be decided efficiently). 
The solution of a linear program representing the given SAT instance provides the basis for six \textit{LP-based features}. 
Finally, there are two sets of so-called \textit{probing features}, which are based on performance statistics over short runs of several well-known SAT algorithms (based on DPLL and stochastic local search, two prominent approaches to solving SAT) and capture the degree to which these make early progress on the given instance.

Some of the feature sets -- specifically, the CG, VG and LP-based features, as well as some of the VCG, balance and DPLL-probing features  -- are computationally quite expensive \citeg{xu2008,hutter2014} and consequently not always useful in the context of practical algorithm selection approaches.
Similarly, the algorithm runs for probing features are limited to a very small part of the overall time budget for solving a given instance, to make sure that sufficient time remains available for running the selected SAT solver.

A decade later, \cite{hutter2014} -- building on the work by \cite{nudelman2004} -- introduced a set of 138 SAT features.
While they removed some features from the earlier sets, much of the set remained the same.
The most significant changes were an extension of the CG and VG feature sets by five new features each, as well as three new feature sets accounting for an additional 48 features. 
The VG feature set was extended by so-called \textit{diameter} features, which capture statistics based on the set of longest shortest paths from one variable to any other one in the graph. 
Also, instead of the weighted clustering coefficients based on the CG \citep[as done by][]{nudelman2004}, \cite{hutter2014} used a set of \textit{clustering coefficients} that measure the CG's ``local cliqueness''. 
Furthermore, they introduced 18 novel \textit{clause learning} features, which summarise information gathered during short runs of a prominent SAT solver, \textsc{zchaff\_rand}, that learns conflict clauses during its search for a satisfying assignment \citep{mahajan2004}.
Another 18 features are derived from estimates of variable bias obtained from the SAT solver \textsc{VARSAT} \citep{hsu2009}; these features essentially capture statistics over estimates for the probability for variables to be $true$, $false$ or $unconstrained$ in every satisfying assignment. 
Finally, \cite{hutter2014} proposed to use the actual feature costs, in terms of the running time required for computing each of the 11 feature sets; they noted that the diameter and survey propagation features tend to be expensive to compute and may thus be of limited usefulness in the context of per-instance algorithm selection.

A well-known generalisation of SAT is the problem of \textit{answer set programming} \citep[ASP; see, \eg{},][]{baral2003}, which deals with determining so-called ``answer sets'', \ie{}, stable models for logic programs. Many combinatorial problems can be presented in ASP in a rather straightforward way and solved, at least in principle, using general-purpose ASP solvers.
Because of the close relationship between ASP and SAT, many features for ASP instances are closely related to the SAT features outlined above.
One of the most widely used collection of ASP features has been proposed by \cite{maratea2012}; it is comprised of 52 features, which can be grouped into four sets.
Three of these feature sets closely correspond to well-known SAT features \citep{nudelman2004} and contain eight \textit{problem size}, three \textit{balance} and two \textit{proximity to Horn} features. 
In addition, \cite{maratea2012} proposed 39 ASP-specific features, such as the numbers of true and disjunctive facts, the fraction of normal rules and constraints, and several combinations of the latter.

\paragraph{AI planning.}

\textit{Automated planning} (also known as \textit{AI planning}) is one of the most prominent challenges in artificial intelligence \citeg{ghallab2004}.
While there are many variants of AI planning problems, the basic setting (also known as classical planning)
involves a set of actions
with associated pre-conditions, deterministic effects and sometimes costs, an initial state and one or more goal states.
The objective in satisficing planning is to find a valid plan, \ie{}, a sequence or partially ordered set of actions that, when applied to the initial state, reach a goal state, or to determine that no valid plan exists.
In the optimisation variants of planning problems, the objective is to find plans of minimal length or cost.
Most variants of AI planning are at least NP-hard, and satisficing classical planning is known to be PSPACE-complete.
AI planning algorithms have important applications, \eg{}, 
in robotics, gaming, logistics and software test case generation;
they are also used for the operation and management of traffic, energy grids and fleets of shared vehicles.

In classical planning, there is an important distinction between a problem instance and a so-called \emph{planning domain}. This distinction arises from the fact that states and actions are specified in an abstract way, using so-called \textit{predicates} and \textit{operators} that can be instantiated to yield specific properties of \textit{states} and specific \textit{actions}, respectively \citeg{ghallab2004}. For example, in a planning problem that involves moving goods using a fleet of trucks, there might be a predicate stating that a specific truck is in a given location, and an operator that moves the truck from one location to another.
A planning domain is a class of a planning instances with the same set of specific predicates and operators.
Planning domains and instances can be concisely described in a widely used, uniform language called \textsc{PDDL} \citep[Planning Domain Definition Language; see, e.g.,][]{gerevini2005}.

\cite{howe1999} were among the first to characterise AI planning instances by simple features, namely, 
the number of actions, predicates, objects, goals, as well as the number of predicates used to specify the initial state.
A decade later, \cite{roberts2008} introduced a substantially extended set of 41 features, which includes summary statistics of the domain and instance files (16 and three features, respectively), but also captures 13 high-level features of the given planning instance in terms of its \textsc{PDDL} requirements. 
They also considered nine features based on the so-called \textit{causal graph (CG)} (\ie{}, a graph capturing causal dependencies between states), such as the number of vertices in the CG and their average degree, as well as various metrics computed from the edges of the CG.

Also based on the idea of using graph properties, \cite{cenamor2013} proposed a total of 47 features, which capture the information contained in causal and \textit{domain transition graphs}. 
The latter represent the permissible transitions between states. 
The causal graph features of \cite{cenamor2013} can be categorised into four different sets: 
(i)~four general graph properties, (ii) four features based on various ratios of graph properties, (iii) 12 statistical aggregations over the entire graph, and (iv) six additional, high-level statistics for states with defined values in the goal specifications. 
The remaining 21 domain transition graph features include (i) three general graph properties (number of edges and states, sum of edge weights) and (ii) 18 statistical features similar to those of the causal graph.

The most recent and extensive collection of AI planning instance features was prepared by \cite{fawcett2014}. 
It contains 12 sets with 311 features in total, covering most of the features from earlier work, as well as a broad range of new ones.
The first three sets extend the 16 domain, three problem and 13 language requirement features from \cite{roberts2008} by two, four and 11 new features, respectively. 
Four further feature sets are based on a translation of the given PDDL instance into a \textit{finite domain representation} (FDR), by means of a well-known AI planning system, \textsc{Fast Downward} \citep{helmert2006}. 
This FDR representation, as well as information collected during the translation and preprocessing, gives rise to sets of 19, 19 and eight features, respectively. 
Building on the work of \cite{cenamor2013}, \cite{fawcett2014} also provide a set of 41 causal and domain transition graph features. 
Six features are computed from information gathered during the preprocessing phases of \textsc{LPG-td} \citep{gerevini2003}, another well-known planning system, while 10 further features capture information produced by the \textsc{Torchlight} local search analysis tool \citep{hoffmann2011};
another 16 features are determined based on the trajectories of one-second probing runs of \textsc{Fast Downward}.
Furthermore, the 115 SAT features from \cite{xu2012} are included, based on a SAT representation of the given planning instance (in form of a CNF with a planning horizon of 10). 
The final feature set introduced by \cite{fawcett2014} contains information on whether the previously outlined sets were computed successfully and additionally captures the respective computation times.

\paragraph{Travelling salesperson problem.}

The \textit{travelling salesperson problem (TSP)} is one of the most intensely studied combinatorial optimisation problems. For decades, it has been the subject of a large body of work and continues to be highly relevant for theoretical analyses, design of algorithms and practical applications ranging from logistics to manufacturing \citeg{applegate2007}.
In the TSP, given an edge-weighted graph, whose vertices are often called \textit{cities} and whose edges represent the cost of travelling from one city to another, the objective is to find a Hamiltonian cycle with minimum total weight, \ie{}, a minimum-cost trip that passes through every city exactly once.
Most work on the TSP focusses on the special case of the two-dimensional Euclidean TSP, where cities are locations in the Euclidean plane, and costs correspond to the Euclidean distances between cities.

The development of features for TSP instances has been initiated by \cite{smithmiles2011}, who proposed the following features for characterising a given TSP instance: (1) coordinates of the instance's centroid, (2) average distance from all cities to the centroid, (3 \& 4) standard deviation of distances, as well as fraction of distinct distances within distance matrix, (5) size of the rectangle enclosing the instance's cities, (6 \& 7) standard deviation, as well as coefficient of variation of the normalised nearest neighbour distances, (8) ratio of number of clusters found by GDBSCAN \citep{sander1998} to the number of all cities, (9) variance of number of cities per cluster, (10) ratio between number of outliers and all cities, and (11) the number of cities. 
Furthermore, \cite{kanda2011} and \cite{kovarik2012} proposed features derived from the distance matrix of a given TSP instance.

Nearly all features from these earlier studies were combined by \cite{mersmann2013} and further extended, leading to a collection of six TSP feature sets with a total of 68 features, many of which are derived from the distance matrix and from the spatial distribution of the cities in the Euclidean plane. 
More precisely, the distance matrix is condensed into \textit{distance} and \textit{mode features}, 
and the distribution of cities is captured by a set of \textit{cluster features}, based on multiple runs of GDBSCAN, as well as \textit{convex hull features}, which quantify the spread of the cities in the Euclidean plane. 
The closeness of neighbouring cities is measured by various \textit{nearest neighbour statistics}, and a final set of features is comprised of the depth and edge costs of the \textit{minimum spanning tree} for the given TSP instance.

\cite{hutter2014} developed a set of 64 TSP features that includes some of the previously outlined instance characteristics as well as new \textit{probing features}.
The latter are based on 20 short runs of a well-known local search solver \citep[LK;][]{lin1973}, as well as single short runs of the state-of-the-art exact TSP solver, Concorde~\citep{applegate2007}.
Probing features were also used by \cite{kotthoff2015}.

The most comprehensive collection of TSP features was provided by \cite{pihera2014}. Their set of 287 features builds on the earlier work by \cite{hutter2014}, but additionally includes instance characteristics derived from the distances of the cities to the convex hull, the number of intersections of locally optimal tours, and statistics of disjoint tour segments. 
The largest group of new features is based on strongly and weakly connected components of so-called $k$-\textit{nearest neighbour graphs}, for many values of $k$.

Finally, we see significant potential for new features based on recent work on funnel-structures in the search landscapes associated with TSP instances \citep{ochoa2015, ochoa2016}.
Considering that highly related aspects of global search space structure have shown to play an important role in algorithm selection for continuous optimisation problems \citep{bischl2012,kerschke2018bbob}, 
features of this nature may also prove to be useful for discrete optimisation problems, such as the TSP.

\paragraph{Other combinatorial problems.}

There is a sizeable body of work on instance characteristics for other combinatorial problems, including the epistasis measures by \cite{davidor1991} and \cite{fonlupt1998}, indicators for the hardness of quadratic assignment \citep[QAP,][]{angel2002} or constraint satisfaction problems \citep[CSP,][]{boukeas2004}, features for so-called orienteering problems, which generalise the TSP by distinguishing between static and dynamic locations \citep{bossek2018local}, and variable interaction measures for combinatorial optimisation problems \citep{seo2007}. 
A detailed discussion of these problems and the respective instance features (some of which are quite generic and can be applied to a range of discrete combinatorial problems) is beyond the scope of this survey;
however, we note that these features, like the problem-specific features described earlier in this section, provide a good basis for per-instance algorithm selection and related tasks.

%%%

\subsection{Continuous problems}
\label{sec:feats_continuous}

We now turn our attention to the optimisation of continuous \textit{fitness landscapes}~\citep{wright1932, kauffman1993}.
In contrast to discrete optimisation problems, which differ very substantially from each other (consider, for example, SAT vs.~TSP) and require problem-specific features for characterising instances, the general idea of continuous optimisation problems can be expressed uniformly, in a rather straightforward way: 
(w.l.o.g.) find the global minimum of an objective or fitness function $f: \mathcal{X} \rightarrow \mathcal{Y}$, which maps vectors of variables, $\mathbf{x} = (x_1, \ldots, x_d)$, from a $d$-dimensional decision space $\mathcal{X} \subseteq \mathbb{R}^d$ (whose values may be subject to additional constraints) to $p$-dimensional objective or fitness values $\mathbf{y} = (y_1, \ldots, y_p) := f(\mathbf{x}) \in \mathcal{Y} \subseteq \mathbb{R}^p$~\citep{jones1995_phd, stadler2002}.
Unfortunately, in most real-world scenarios, the exact mathematical representation of the fitness function $f$ is unknown. Thus, its optimisation often has to be handled as a black-box problem, and consequently becomes difficult and expensive. 
In light of this, it is especially useful to characterise a specific problem by means of (informative) features, based on which it is possible to select a suitable optimisation algorithm.

As there only exist preliminary studies on the characterisation of multi-objective ($p \ge 2$) problems~\citep[see, e.g.,][]{kerschke2016_ppsn} -- which we will discuss later -- we will in the following mainly focus on the manifold of characterisation approaches for single-objective ($p = 1$) continuous optimisation problems.

\paragraph{Single-objective continuous problems.}

Overviews on the early works related to this problem class can be found in \cite{pitzer2012}, \cite{malan2013}, \cite{sun2014} and \cite{munoz2015_as}. 
However, in contrast to recent studies, the majority of the studies covered by those surveys proposed measures for characterising white-box problems -- i.e., problems, whose landscapes are entirely known upfront -- rather than cheap, informative and automatically computable landscape features as needed for black-box problems. 
Obviously, only the latter are beneficial for automated algorithm selection (or related problems). 
Nevertheless, we briefly discuss some noteworthy contributions to the characterisation of white-box problems, as they form the basis for recent developments.

In the 1990s, landscapes were classified into easy and hard problems -- from an optimisation algorithm's point of view. While \cite{jones1995} \citep[and later][]{mueller2011} proposed \textit{fitness distance correlation} as a key characteristic, \cite{rose1996} suggested the \textit{density of states} for solving the binary classification task. 
Furthermore, so-called \textit{epistasis} measures \citep{naudts1997, rochet1997} quantify the influence of single variables (or bits, in case of a bit-representation of $\mathbf{x}$) on the problem's fitness, which in turn can be used to rank the landscapes according to their difficulty. 
The \textit{information content} measures from \cite{vassilev2000} provide another basis for quantifying hardness.

In the following decade, attention shifted from characterising problem difficulty to \textit{ruggedness}. 
Depending on \textit{fitness evolvability portraits} \citep{smith2002}, \textit{autocorrelation coefficients}, number and \textit{distribution of optima} \citep{brooks2003}, or \textit{entropy} \citep{malan2009}, landscapes were categorised into rugged, neutral and smooth problems. 
During that time, researchers also focused on problem multimodality \citep{preuss2015}, i.e., the analysis of the problems' landscapes w.r.t.~multiple local and/or global optima. 
For instance, the \textit{barrier trees} of \cite{flamm2002} provided means for identifying basins of attractions, local optima and saddle points within the landscapes, and the \textit{dispersion} metrics from \cite{lunacek2006} enabled an approximate estimation of the degree of multimodality of a given landscape.

\begin{figure}[!t]
\centering
\includegraphics[width=\textwidth]{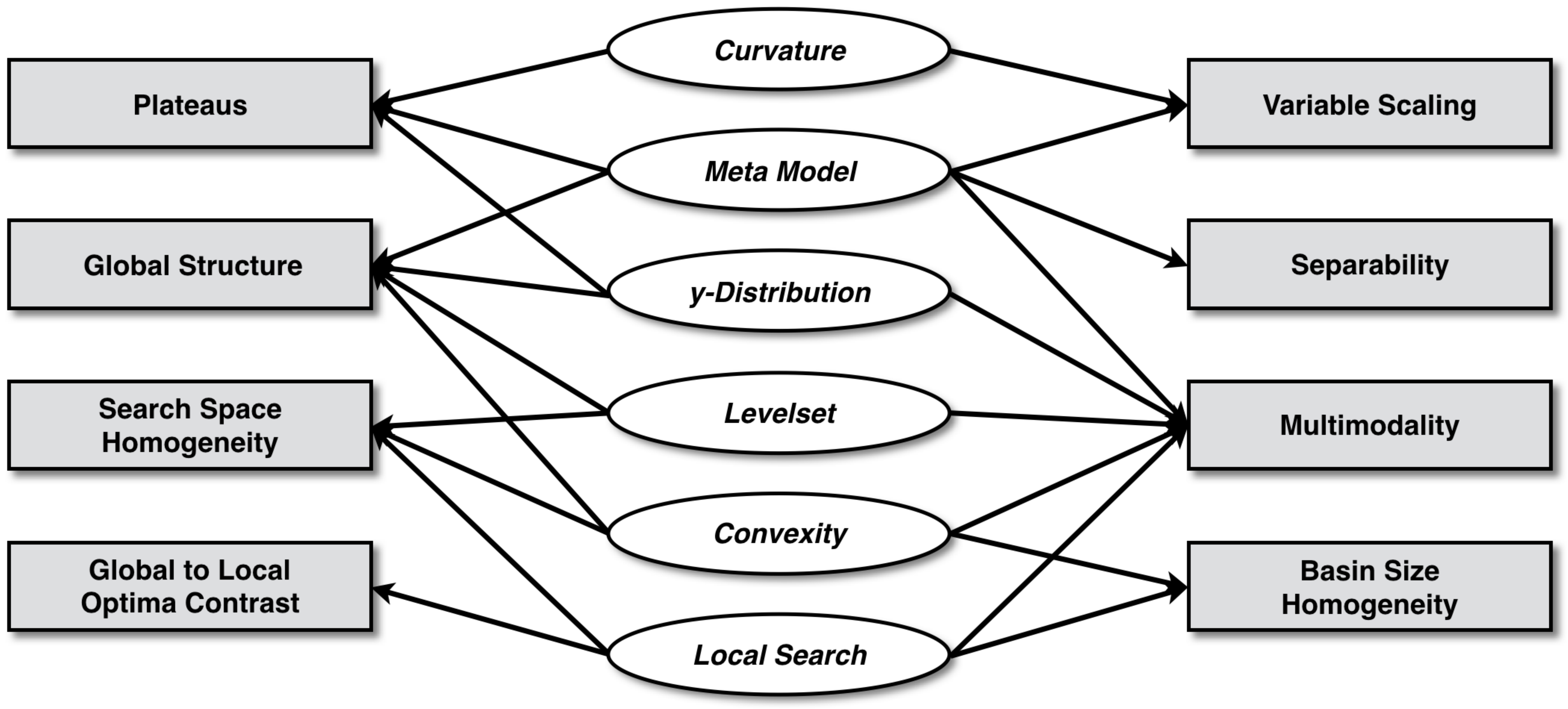}
\caption{Overview of connections between ``high-level'' properties (grey rectangles) and ``low-level'' features (white ellipses), taken from \cite{mersmann2011}.}
\label{fig:ela}
\end{figure}

\cite{pitzer2012} combined these approaches under the term \textit{fitness landscape analysis (FLA)}. 
However, as the majority of those characteristics was proposed for white-box settings, they are not useful for automated algorithm selection. 
With the introduction of \textit{Exploratory Landscape Analysis (ELA)}, \cite{mersmann2011} were the first to explicitly develop landscape features for use in black-box optimisation (BBO) -- and hence for algorithm selection in continuous optimisation. 
As shown in Figure~\ref{fig:ela}, they introduced six feature sets (\textit{curvature}, \textit{convexity}, \textit{levelset}, \textit{local search}, \textit{meta models} and $y$-\textit{distribution}), combining a total of 50 features, and used these to predict eight different problem characteristics, such as the global structure (none, weak, strong, deceptive) and the degree of multimodality (none, low, mediocre, high). 
We note that the attributes of the latter ``high-level'' properties can only be assigned by someone with knowledge of the entire problem, whereas the former ``low-level'' features can be automatically computed based on a small, but representative sample of points -- the so-called \textit{initial design}.

Following the idea of automatically computable numerical features, some of the earlier white-box landscape characteristics have been adapted to the black-box context. 
For instance, \cite{munoz2012, munoz2015_ic} refined the \textit{information content features} and extended them by \textit{basin of attraction features}. 
Similarly, \cite{abell2013} proposed \textit{hill climbing} and \textit{random point features}, whose general idea stems from the local search and fitness distance correlation methods, respectively, and enhanced them by \textit{problem definition measures}.

In addition, several new feature sets have been introduced in recent years: \cite{kerschke2014} discretised the continuous search space into a grid of cells and used a Markov-chain-inspired cell-mapping approach to obtain features, which measure -- amongst others -- the sizes of the basins of attraction. 
However, due to the curse of dimensionality, those features are only practically applicable to low-dimensional problems. 
A much more scalable measure for the basins of attraction and the global structure of a landscape are the \textit{nearest better clustering features} \citep{kerschke2015}, which provide the means to distinguish funnel-shaped from rather random global structures. 
The most recently published features based on aggregated information of neighbouring points are the \textit{bag of local landscape features} by \cite{shirakawa2016}. 
Furthermore, the \textit{length scale features} of \cite{morgan2015} measure the variable scaling based on the ratio between the change in objective space and distance in decision space for pairs of distinct observations from the initial design.

We note that work in this area is not restricted to the development of features or problem characteristics. 
For instance, \cite{malan2015} and \cite{bagheri2017} investigated constrained optimisation problems, whereas \cite{kerschke2016_budget} showed that landscape features already possess sufficient information if they are computed based on rather small samples of $50 \cdot d$ observations (where $d$ is the dimensionality of the given optimisation problem).

Until recently, the simultaneous use of feature sets from different research groups has been cumbersome and hence rarely practiced. 
However, with the development of \href{https://cran.r-project.org/package=flacco}{\texttt{flacco}}~\citep{kerschke2017_flacco,kerschke2017flaccoPkg}, an \texttt{R}-package that provides source code for most of the previously listed ELA features, this obstacle has been overcome. 
Since then, the complementarity of the various features and their potential usefulness as a basis for algorithm selection has been demonstrated in several studies. 
We provide a detailed overview of this work later (see Section~\ref{sec:continuous}). Note that by using a platform-independent web-application of the \texttt{flacco} package\footnote{\url{https://flacco.shinyapps.io/flacco/}}~\citep{hanster2017}, researchers and practitioners, who are unfamiliar with \texttt{R}, can also benefit from this extensive collection of more than 300 landscape features.
\cite{belkhir2016,belkhir2017} were among the first to leverage the ELA features provided by \texttt{flacco} for per-instance algorithm configuration.

\paragraph{Multi-objective continuous problems.}

While the characterisation of single-objective continuous optimisation problems has been studied for over two decades, only preliminary studies have been conducted w.r.t.~informative features of multi-objective problems. 
For instance, \cite{kerschke2016_flacco} used features that have been developed for single-objective problems and used them to cluster some well-known and frequently used multi-objective benchmark problems. 
However, those features do not capture characteristics that are especially important to multi-objective problems, such as interaction effects between the objectives. 
Eventually, techniques that were originally aimed at measuring variable interactions \citep[see, e.g.,][]{reshef2011,sun2017} could help to overcome these limitations.

\begin{figure}[!t]
\centering
\includegraphics[width=\textwidth, trim = 13mm 25mm 13mm 10mm, clip]{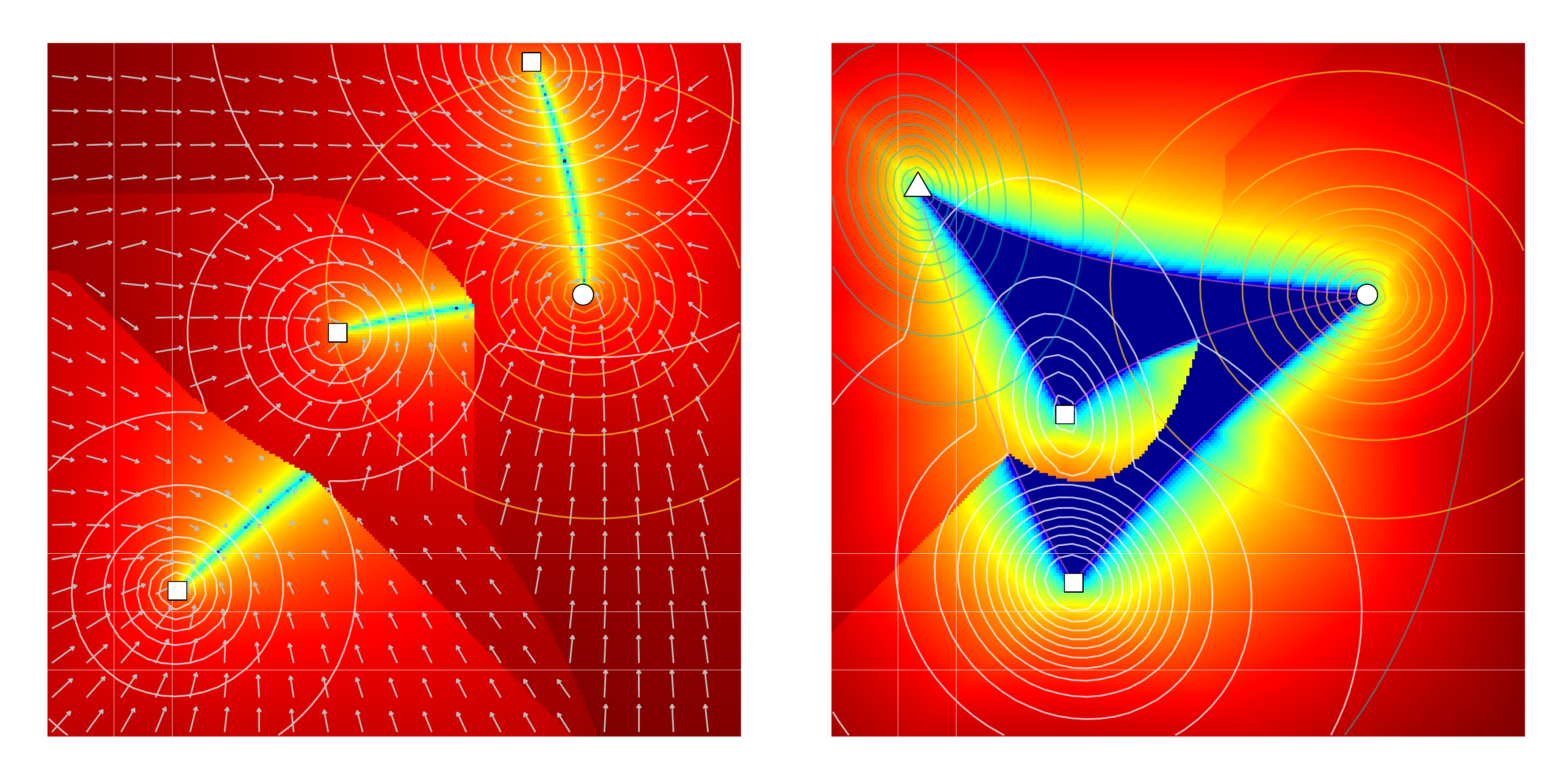}
\caption{Exemplary visualisations of the \textit{gradient field heatmaps} proposed by \cite{kerschke2017_emo}. The images show the decision spaces of continuous optimisation problems with two (left) and three objectives (right), respectively. As a result of the interactions between the local optima of the different objectives (indicated by circles, squares and triangles), the landscapes show multiple basins of attraction.}
\label{fig:gradfield_heatmap}
\end{figure}

\cite{kerschke2016_ppsn,kerschke2018search} investigated locally efficient sets and the corresponding locally optimal fronts (\ie{}, the multi-objective equivalents of local optima) and proposed measures -- on the basis of those sets and fronts -- which enable the distinction of multi-objective problems according to their degree of multimodality. 
Interestingly, those studies revealed that even in the case of rather simple multi-objective problems, strong interaction effects between the objectives exist. 
Thus, in order to improve the understanding of multi-objective problems, \cite{kerschke2017_emo} introduced new techniques for visualising them. 
Their approach,  dubbed \textit{gradient field heatmaps} (see, \eg{}, Figure~\ref{fig:gradfield_heatmap}), is based on visualising the decision space, but nevertheless clearly reveals interactions between the objectives -- in the form of rugged landscapes with several basins of attractions. 
Unexpected findings of this kind \citeg{grimme2018} indicate that research in this area is still at an early stage, leaving many open challenges for future work (see Section~\ref{sec:perspectives} for further details).

\section{Algorithm Selection for Discrete Problems}
\label{sec:discrete}

Over the years, algorithm selection techniques have achieved remarkable results in several research areas -- especially for discrete combinatorial problems \citeg{smithmiles2009,kotthoff2014}. 
However, due to the significant differences between various problems, not only the respective instance features, but also solvers and algorithm selectors vary considerably. 
Therefore, it is impossible to cover all work in this area; instead, in the following, we focus on a small number of particularly well-known problems: \textit{propositional satisfiability (SAT)} and related problems, \textit{AI planning} and the \textit{travelling salesperson problem (TSP)}.

%%%

\paragraph{Propositional satisfiability and related problems.}

Historically, some of the first and most widely known successes of per-instance algorithm selection have been achieved in the context of solving the \textit{propositional satisfiability problem (SAT)}. 
We note that several approaches described in the following can be applied to closely related problems (such as MaxSAT) in a rather straightforward way. 

\subparagraph{SATzilla2003 and SATzilla2007.} Within the highly contested area of SAT, the first AS system to outperform stand-alone SAT solvers was \textsc{SATzilla}. While its first version, denoted \textsc{SAT\-zilla\-2003} \citep{nudelman2004satzilla}, still showed (minor) weaknesses -- it ``only'' ranked second (twice) and third in the 2003 SAT Competitions -- its (enhanced) successor, \textsc{SATzilla2007} \citep{xu2008}, won multiple prizes (3x first, 2x second and 2x third) in the 2007 SAT Competition. 
Despite the differences w.r.t.~their success, the general working principles underlying both systems are quite similar. During the actual construction phase, pre- and backup solvers are identified, based on the performances of the given solvers on training data. 
All instances that have not been solved by the respective pre-solvers are then used for training separate regression models per solver, which in turn are used for selecting a promising subset of ``main solvers''.

The main difference between \textsc{SATzilla2003} and \textsc{SATzilla2007} lies in the regression models used for the main algorithm selection phase: 
While the former used \textit{empirical hardness models} \citep{leytonbrown2002} based on \textit{ridge regression} \citeg{bishop2006}, the latter employed \textit{hierarchical hardness models} \citep{xu2007}, more precisely \textit{sparse multinomial logistic regression} \citep[SMLR, see, \eg{},][]{krishnapuram2005}. 
The latter version of \textsc{SATzilla} sequentially runs up to two manually selected pre-solvers; if these fail to solve the given instance within a user-specified running time budget, instance features are computed and used to predict (and sequentially run) the best solver(s) from the given set $\mathbf{A}$. 
The system either terminates once the given SAT instance has been solved, or when a user-specified cutoff time is reached.

\cite{xu2008} compared different versions of \textsc{SATzilla2007} on 2\,300 random (RAND), 1\,490 crafted/handmade (HAND) and 1\,021 application/industrial (INDU) instances, using the same setup as the 2007 SAT Competition. 
For this purpose, they used a total of 48 features and identified between one and two pre-solvers, one backup solver, as well as three to five main solvers from a given set $\mathbf{A}$ (which differed between RAND, HAND and INDU).
Notably, their pre-solvers solved between 32 and 62\% of the instances -- despite running for only seven CPU seconds at most. 
Ultimately, \textsc{SATzilla2007} ranked first (inter alia) in the SAT \& UNSAT (RAND and HAND) categories of the competition, as well as in UNSAT (HAND).

\subparagraph{3S.} \cite{kadioglu2011} proposed a hybrid approach, denoted \textit{semi-static solver schedules} (\textsc{3S}), which combines algorithm selection with solver scheduling. 
Since it can be very expensive to determine schedules over all solvers from a given set, \cite{kadioglu2011} devised a different approach, in which they partitioned the (normalised) instance feature space for a given training set by means of $g$-means \citep{hamerly2003}. 
Then, the best value of $k$ to be used by the $k$-nearest neighbour algorithm \citep{hastie2009} was identified for each cluster and directly used for training semi-static solver schedules. 
\textsc{3S} was demonstrated to perform very well; \eg{}, it reduced the gap between the (back then) state-of-the-art selector on RAND instances, \textsc{SATzilla2009\_Rand}, and the VBS by 57\%. 
Additionally, this general-purpose selector performed very well at the 2011 SAT Competition, winning seven medals (including two gold) without training separate selectors for different competition tracks (as had been done for previous versions of \textsc{SATzilla}).

\subparagraph{SATzilla2012.} Building on the success of earlier versions of \textsc{SATzilla}, \cite{xu2012} developed \textsc{SATzilla2012}, which showed outstanding performance in multiple SAT competitions. 
\textsc{SATzilla2012} uses cost-sensitive pairwise classification as the basis for per-instance algorithm selection; these penalise incorrect predictions according to the loss in performance caused by them.
More precisely, one cost-sensitive decision tree \citep{ting2002} is used for every pair of solvers in the given set $\mathbf{A}$ to predict which of the two solvers will perform better on the given instance.
Simple voting over these pairwise predictions is used to determine the solver to be run. 
Like earlier versions of \textsc{SATzilla}, \textsc{SATzilla2012} uses pre- and backup-solvers in addition to the main algorithm selection stage. Additionally, a decision forest \citep{hastie2009} based on the number of clauses and variables in the given SAT instance is used to predict whether the computation of further instance features is sufficiently cheap to proceed to the main selection stage (otherwise, a statically chosen backup solver is run).
In the 2012 SAT Competition, \textsc{SATzilla2012} used a total of 31 SAT solvers and 138 features and ended up winning multiple prizes.

\subparagraph{CSHC.} Building on the static scheduling approach underlying \textsc{3S}, \cite{malitsky2013} introduced an algorithm selection system based on the core concept of \textit{cost-sensitive hierarchical clustering} (\textsc{CSHC}). 
During its training phase, \textsc{CSHC} iteratively partitions the instance feature space by means of hyperplanes, and occasionally undoes splits if that leads to improvements in overall performance. 
When given a new instance, \textsc{CSHC} first runs the \textsc{3S} static algorithm schedule for 10\% of the overall running time allotted for solving a given SAT instance $F$ and -- in case $F$ remains unsolved -- runs the SAT solver that performed best on the partition to which $F$ belongs.
The resulting CSCH algorithm selection system has been reported to achieve even better performances than \textsc{SATzilla2012} on a slightly modified version of the 2011 SAT competition \citep[for details, see][]{malitsky2013}.

\subparagraph{SNNAP.} An approach dubbed \textit{solver-based nearest neigbour for algorithm portfolios} \citep[\textsc{SNNAP};][]{collautti2013} successfully combines clustering with per-instance algorithm selection. 
It uses \textit{random forests} \citep{hastie2009} to predict the running times of individual solvers. 
However, instead of directly selecting a solver based on these predictions, SNNAP uses them to identify SAT instances from the given training set that are similar to the instance to be solved.
Specifically, instance similarity is quantified by means of the \textit{Jaccard distance} -- whose distance between two sets $A$ and $B$ is defined as $d(A, B) = 1 - |A \cap B| / |A \cup B|$ -- applied to binary vectors indicating a (small) fixed number of best solvers per instance. 
\textsc{SNNAP} then selects the solver that performed best on the $k$ nearest neighbours of the given instance, where $k$ is a user-defined constant.
According to results reported by \cite{collautti2013}, despite the simplicity of the approach, \textsc{SNNAP} closes around 50\% of the VBS-SBS gap on a broad set of well-known SAT instances.

\subparagraph{AutoFolio.} In light of the many design choices encountered in the development of state-of-the-art algorithm selection systems, \cite{lindauer2015,lindauer2017autofolio} proposed a powerful combination of per-instance algorithm selection and automated algorithm configuration: \textsc{AutoFolio}. 
In a nutshell, \textsc{AutoFolio} applies the automated algorithm configurator \textsc{SMAC} \citep{hutter2011} to the highly parametric algorithm selection framework \textsc{claspfolio2} \citep{hoos2014}. The space is structured in layers, starting with parameters for pre-solving schedules (including their allocated budgets), pre-processing procedures (transformations, filtering, etc.) and the algorithm selection systems (resembling a broad range of approaches, including \textsc{SATzilla2007}, \textsc{SATzilla2012}, \textsc{3S} and \textsc{SNNAP}). 
Subsequent layers specify additional design decisions and (hyper-)parameters. 
As demonstrated for multiple algorithm selection scenarios from the ASLib, \textsc{AutoFolio} indeed achieves results that are highly competitive with those of the best-performing selection systems for a broad range of algorithm selection scenarios, without the need for manual choice of the selection mechanism or selector parameters \citep{lindauer2015,lindauer2017autofolio}.

%%%

\paragraph{AI planning.}

The notion of algorithm selection can be applied to domains as well as to instances of AI planning problems.
In the first case, a planner is selected for a specific domain and then applied for solving arbitrary instances in that domain.
This is conceptually closely related to per-set algorithm selection, as discussed in Section~\ref{sec:related}.
In the second case, a planner is selected for a specific instance of a planning problem, such that even within the same domain, different planners may be chosen, depending on the characteristics of the specific problem instance.
Unlike per-domain selection of planners, this is an instance of per-instance algorithm selection and hence will be our focus in the following.

\subparagraph{Per-domain selection approaches.} 
Because per-domain selection of AI planners has been prominently studied in the literature, we briefly discuss some well-known approaches. 
The \textsc{PbP} planning system and its successor, \textsc{PbP2}, are based on the idea of statistically analysing the performance of several domain-independent planning algorithms on a set of training instances from a given planning domain in order to select a set of planners and associated running times \citep{gerevini2009,gerevini2011}. When solving new instances from the same domain, these planners are run one after the other, using round-robin scheduling with the pre-determined running times for each planner.  \textsc{PbP} and \textsc{PbP2} also make use of macro-actions, sequences of actions whose judicious use can considerably improve planner performance. 
\textsc{PbP} was the overall winner of the learning track of the 6th International Planning Competition, and \textsc{PbP2} brought further improvement through the integration of automated algorithm configuration to better exploit the performance potential of parameterised planners.

The \textsc{ASAP} planning system is based on similar ideas \citep{vallati2013, vallati2014}. In addition to macro-actions, \textsc{ASAP} also exploits so-called entanglements,  which reflect causal relationships that are characteristic for a given domain of planning problems. Different from \textsc{PbP} and \textsc{PbP2}, \textsc{ASAP} selects only a single representation (set of macro-actions and entanglements) and planner for a given domain. On standard benchmarks, \textsc{ASAP} has been demonstrated to outperform \textsc{PbP2} (as well as all the component planners it uses) in terms of the quality or cost of the plans found \citep{vallati2014}.

\subparagraph{IBaCoP2.}
To the best of our knowledge, the first successful application of per-instance algorithm selection to AI planning
was demonstrated by \textsc{IBaCoP2} \citep{cenamor2014}.
\textsc{IBaCoP2} uses 12 component planners, which were selected based on their performance on a large and diverse set of problem instances from past international planning competitions, applying Pareto efficiency analysis to the solution quality of the best plan found by a given planner, and the time required to find the first valid plan.
A random forest model \citeg{hastie2009}, learned from performance data of the component planners on a set of training instances using WEKA \citep{witten2016}, forms the core of the algorithm selection strategy.
This model is used to predict whether a component planner will solve a given problem instance within a fixed time limit, based on a set of 35 cheaply computable, domain-specific features, some of which are derived from heuristics used in state-of-the-art planning algorithms.
To hedge against the consequences of poor predictions, \textsc{IBaCoP2} selects the five component planners with the highest estimated probability $p$ for solving the given instance $i$ and runs these, in the order of decreasing values of $p$, one after the other, each for one fifth of the overall time given to the selector for solving $i$.
We note that, because of this latter strategy, \textsc{IBaCoP2} is not a pure per-instance algorithm selector, but rather combines per-instance algorithm selection with a simple algorithm scheduling approach (\emph{cf.} Section~\ref{sec:related}).
\textsc{IBaCoP} and \textsc{IBaCoP2} showed strong performance in the 8th International Planning Competition, with \textsc{IBaCoP2} winning the sequential satisficing track \citep{vallati2015}.

\subparagraph{Planzilla.}
A second per-instance algorithm selection approach for AI planning has been considered by \citet{rizzini2015,rizzini2017}. 
Their \textsc{Planzilla} system can be seen as an application of the previously outlined \textsc{$^{\ast}$zilla} approach \citep{cameron2017oasc} to AI planning.
Based on the default configuration of \textsc{$^{\ast}$zilla}, \textsc{Planzilla} is comprised of four sequential stages:
(1) a static pre-solving schedule, (2) feature computation, (3) per-instance algorithm selection and (4) a backup solver.
The pre-solving schedule is obtained by greedy selection from the given set of component planners and allocated 1/90 of the overall time budget for solving the given instance $i$. 
Training instances solved during the pre-solving stage are not considered for constructing the per-instance selector, nor for selecting the backup solver.
Per-instance algorithm selection makes use of a comprehensive set of 311 features that includes a broad range of properties of instance $i$, as well as features derived from encoding $i$ into propositional satisfiability.
Based on this set of features, \textsc{Planzilla} uses cost-sensitive classification forests for each pair of component planners in combination with a voting procedure to determine the planner to be run for the remainder of the given time budget.
Before computing the complete feature set, which can be somewhat costly, \textsc{Planzilla} uses a simple model to predict whether feature computation can be completed within the remaining time, $t'$; if not, feature computation and per-instance algorithm selection are skipped, and a backup solver is run instead. 
This backup solver is also run if the component planner selected in stage 3 terminates early without producing a valid plan; it is determined as the solver with the best performance for running time $t'$ on the set of problem instances used to train \textsc{Planzilla} (excluding any instances solved during pre-solving).

Using all planners that participated in the optimal track of the 2014 International Planning Competition (IPC-14),
\textsc{Planzilla} was found to substantially outperform these individual planners and achieve performance close to that of the VBS \citep{rizzini2015,rizzini2017}.
However, when evaluated on a set of testing instances dissimilar from those used for training, it was found that dynamic algorithm scheduling approaches performed better than \textsc{Planzilla}; these approaches dynamically construct an algorithm schedule by performing multiple stages of per-instance algorithm selection, using not only features of the planning instance $i$ to be solved, but also taking into account which component planners have already been run on $i$, without success, in earlier stages of the schedule.
 
%%%

\paragraph{Travelling salesperson problem.}

The potential for per-instance algorithm selection for the TSP differs markedly between \textit{exact TSP solvers}, which are guaranteed to find provably optimal solutions for given TSP instances, and \textit{inexact solvers}, which may find optimal solutions, but cannot produce a proof of optimality.
In exact TSP solving, there is a single algorithm, \textsc{Concorde} \citep{applegate2007}, that has defined state-of-the-art performance for more than a decade.
In contrast, for inexact TSP solving, there is no single algorithm that clearly dominates all others (across all types of instances). 
In fact, several studies~\citep{pihera2014,kotthoff2015,kerschke2017_tsp} have shown that at least three TSP solvers, \textsc{EAX} \citep{nagata1997,nagata2013}, \textsc{LKH} \citep{helsgaun2000,helsgaun2009} and \textsc{MAOS} \citep{xie2009}, define state-of-the-art performance on different kinds of TSP instances.
In addition, two enhanced versions of \textsc{EAX} and \textsc{LKH} (denoted \textsc{EAX+restart} and \textsc{LKH+restart}), which employ additional restart mechanisms to overcome stagnation in the underlying search process, often (but not always) outperform \textsc{EAX} and \textsc{LKH}, respectively \citep{duboislacoste2015}.
Further modifications of these algorithms -- \eg{}, based on the alternative crossover operator proposed by \cite{sanches2017building, sanches2017}, which recently was integrated into LKH \citep{tinos2018} -- might achieve even better performance; however, to this date, we are not aware of conclusive evidence to this effect.
Instead, using automated algorithm selection techniques, the performance complementarity between existing solvers has been leveraged, leading to very substantial performance improvements over the single best solver \citep{kotthoff2015,kerschke2017_tsp}. 

\cite{kotthoff2015} compared \textsc{EAX}, \textsc{LKH} and their respective restart variants across four well-known sets of TSP instances: random uniform Euclidean (RUE) instances, problems from the TSP library (TSPLIB), as well as national and VLSI instances\footnote{\url{http://www.math.uwaterloo.ca/tsp/index.html}}. 
They used the feature sets proposed by \cite{mersmann2013} and \cite{hutter2014} (see Section~\ref{sec:feats_discrete}) for constructing multiple algorithm selectors. 
Their best selector, based on multivariate adaptive regression splines \citep[MARS; see, \eg,][]{friedman1991}, was trained on a pre-defined subset of features by \cite{hutter2014} and closed the gap between the single best solver from their set (\textsc{EAX+restart}) to the VBS by 10\%.

In an extended version of this earlier study, \cite{kerschke2017_tsp} considered additional types of TSP instances, feature sets and solvers, and furthermore employed more sophisticated machine learning techniques, including various feature selection strategies, for constructing algorithm selectors. 
Specifically, the set of TSP instances was extended by clustered and morphed instances \citep{gent1999,mersmann2012,meisel2015}, \ie{}, linear combinations of clustered and RUE instances\footnote{generated using the \textsc{R}-package \textsc{netgen}~\citep{bossek2015}}.
Furthermore, the basis for algorithm selection was expanded with the feature set of \cite{pihera2014} and the \textsc{MAOS} solver by \cite{xie2009}.
The best algorithm selector under this extended setup was found to be a support vector machine \citep{karatzoglou2004}, which was constructed on a cheap, yet informative, subset of 16 nearest-neighbour features \citep{pihera2014}. 
This particular selector achieved -- despite its non-negligible feature computation costs -- a PAR10 score of 16.75s and thereby closed the gap between the single best solver (\textsc{EAX+restart}; 36.30s) and the virtual best solver (10.73s) by more than 75\%.

%%%

\paragraph{Further discrete combinatorial problems.}

Throughout the previous paragraphs, we gave an overview of algorithm selection approaches for some of the most prominent and widely studied discrete combinatorial problems.
Of course, per-instance algorithm selection has shown to be effective on several other discrete problems -- such as the \textit{travelling thief problem (TTP)}, where \cite{wagner2017} recently presented the first study of algorithm selection, along with a comprehensive collection of performance and feature data.

In some cases, successful applications of algorithm selection techniques have been described using different terminology.
For instance, \cite{smith2008} presented her algorithm selector for the \textit{quadratic assignment problem (QAP)} under the umbrella of a ``meta-learning inspired framework''. Similarly, \cite{pulina2009} demonstrated the successful application of algorithm selection to the problem of solving \textit{quantified Boolean formulae (QBF)}, an important generalisation of SAT; yet, they describe their selector as a self-adaptive multi-engine solver, which ``selects among its reasoning engines the one which is more likely to yield optimal results''.

Considering the size of the literature on per-instance algorithm selection for discrete combinatorial problems, a comprehensive overview would be far beyond the scope of this survey and produce little additional insight.
Instead, we will now shift our attention to continuous problems, which present different challenges and opportunities for algorithm selection.

\section{Algorithm Selection for Continuous Problems}
\label{sec:continuous}

As previously mentioned, the efficacy of algorithm selection methods strongly depends on the performance data used for training them:
The more representative the training set is regarding the entire range of possible problem instances, the better performance we can expect on previously unseen instances. 
For continuous optimisation problems, representativeness of benchmark sets has been a matter of long-standing debate, ranging from early works of \cite{dejong1975} up to more recent sets, \eg{}, from the CEC competitions~\citep{li2013} or the black-box optimisation benchmark \citep[BBOB,][]{hansen2009} collection. 
There are several specific function generators, such as a framework for generating test functions with different degrees of multimodality~\citep{wessing2015}. 
Some of the most frequently used, and arguably most relevant test functions are included in the \texttt{Python}-package \texttt{optproblems}~\citep{wessing2016} and the \texttt{R}-package \texttt{smoof}~\citep{bossek2017}. While all these benchmark sets (and generators) have advantages and drawbacks, a detailed discussion is beyond the scope of this article. 
However, we note that the construction of representative training sets remains, at least to some degree, an open challenge in the context of algorithm selection for continuous optimisation problems.

%%%

\paragraph{Unconstrained single-objective optimisation problems.}

In single-objective continuous optimisation, only few studies directly and successfully address the algorithm selection problem in an automated way. 
An initial approach of combining exploratory landscape analysis (ELA) and algorithm selection was presented by \cite{bischl2012}, focusing on the BBOB test suite. 
The latter consists of 24 functions which are grouped into four classes, mainly based on their multimodality, separability and global structure. 
Each function is represented by different instances resulting from slightly varied function parametrisations. 
Within the BBOB competition, 15 algorithm runs had to be conducted per function, equally distributed among five (BBOB 2009) or 15 (BBOB 2010) instances for decision space dimensions 2, 3, 5, 10, 20 and, optionally, 40.  Algorithm performance was then evaluated using expected running time \citep[ERT,][]{hansen2009}, which reflects the expected number of function evaluations required to reach the global optimum up to a threshold of $\varepsilon > 0$.
Subsequently, the ERT was divided by the ERT of the best algorithm for this function within the respective competition to obtain a relative ERT indicator. 
Within the BBOB setup, accuracies in the range of $\{10^{-3},10^{-4},\ldots,10^{-8}\}$ are considered.

A representative set of four optimisation algorithms was constructed from the complete list of candidate solvers.
Based on the low-level features introduced by \cite{mersmann2011}, \cite{bischl2012} aimed for an accurate 
prediction of the best of the four algorithms for each function within the benchmark set. 
For this purpose, a sophisticated cost-sensitive learning approach, based on one-sided support vector regression with a radial basis function kernel, was used \citep{tu2010}.  
This complex approach enabled the minimisation of loss (measured by relative ERT) due to incorrect predictions.
The median relative ERT of all 600 entries (five runs times five instances for each of the 24 BBOB functions)
served as the overall performance measure of the resulting classifier. 
Two different cross-validation strategies were investigated: cross-validation over the instances of each given function or over the complete set of functions. 
The latter task can certainly be considered more challenging, due the structure of the BBOB test set, which was designed to comprise 24 functions with maximally diverse characteristics, covering a broad range of continous optimisation problems. 
While, as expected, better performance was observed in the first setting, performance was remarkably high in both cases.

Recently, substantial progress has been made in terms of systematically constructing an automated algorithm selector for a joint dataset of all available BBOB test suites \citep{kerschke2017_thesis,kerschke2018bbob}, making use of the \texttt{R}-packages \texttt{flacco} \citep{kerschke2017_flacco,kerschke2017flaccoPkg}, \texttt{smoof} \citep{bossek2017} and \texttt{mlr} \citep{bischl2016_mlr}. 
A total of 480 BBOB instances (instance IDs 1--5 of all 24 BBOB functions, across dimensions 2, 3, 5 and 10) were considered, combined with respective results of all 129 solvers submitted  to the COCO platform \citep{hansen2016} so far. 
In order to keep the size of the set of solvers manageable and to focus on the most relevant and high performing solvers, the construction of the algorithm selector was based on a carefully selected subset of 12 solvers: two deterministic methods \citep[variants of the \textsc{Brent-STEP} algorithm, \textsc{BSrr} and \textsc{BSqi}, see][]{baudivs2015}, five multi-level approaches --  \textsc{MLSL} \citep{pal2013, rinnooy1987}, \textsc{fmincon}, \textsc{fminunc}, \textsc{HMLSL} \citep{pal2013} and \textsc{MCS} \citep{huyer2009} -- as well as four \textsc{CMA-ES} variants: \textsc{CMA-CSA} \citep{atamna2015}, \textsc{IPOP-400D} \citep{auger2013}, \textsc{HCMA} \citep{loshchilov2013} and \textsc{SMAC-BBOB} \citep{hutter2013}. 
The commercial solver \textsc{OptQuest/NLP} \citep{pal2013,ugray2007} was also included.

Performance was measured by relative ERT as in \cite{bischl2012} by normalising the ERT for each solver per problem and dimension based on the best ERT for the respective problem (among the algorithms in the given set). 
A hybrid version of \textsc{CMA-ES} \citep[\textsc{HCMA},][]{loshchilov2013} turned out to be the single best solver (SBS), with a mean relative ERT score of 30.4 across all considered instances -- and thus being the only solver to approximate the optimal objective value of all 96 problems up to the precision level of $10^{-2}$ used for this study.
Various combinations of supervised learning (notably, classification, regression, pairwise regression) methods and feature selection strategies (greedy forward-backward and backward-forward, two genetic algorithm variants) were utilised in combination with leave-one-function-out cross-validation. 
The best algorithm selector obtained in this manner, a classification-based support vector machine \citep{vapnik1995} combined with a sequence of a greedy forward-backward and a genetic-algorithm-based feature selection approach, managed to reduce the mean relative ERT of the SBS roughly by half, to a value of 14.2, only requiring nine out of more than 300 exploratory landscape features. 
Specifically, meta-model and nearest-better clustering features (see Section \ref{sec:feats_continuous}) were used in this context. 
Feature computation costs were taken into account and accounted for merely $50 \times d$ samples of the objective function \citep{kerschke2016_budget} -- where $d$ denotes the dimensionality of the given decision space -- which matches common intitial population sizes of evolutionary optimisation algorithms. 
Hence, when using such an evolutionary optimisation algorithm, making use of its initial population -- which needs to be evaluated in any case -- renders feature computation cost negligible.

%%%

\paragraph{Constrained single-objective optimisation problems.}

In the area of constrained continuous optimisation, the performance of popular evolutionary computation techniques -- such as differential evolution, evolution strategies and particle swarm optimisation -- has been investigated for problems with linear and quadratic constraints~\citep{poursoltan2015_comparison,poursoltan2015_analysis}. 
Features capturing the correlation of these constraints have been investigated w.r.t.~their impact on solver performance.
\cite{malan2015} and \cite{malan2018} numerically characterised constraint violations using landscape features.
Based on results on the CEC 2010 benchmark problems, it was demonstrated that this approach produces detailed insights into constraint violation behaviour of continuous optimisation algorithms, indicating its potential usefulness in the context of algorithm selection and related approaches.
	
Furthermore, constraints have been evolved in different ways to construct instances that can be used for algorithm selection. 
This includes approaches maximising the performance difference of two given solvers, as well as a multi-objective approach for creating instances that reflect the tradeoffs between two given algorithms observed when varying the constraints. 
\cite{neumann2016} demonstrated that the multi-objective approach leads to an instance set that provides a better basis for algorithm selection than sets obtained by maximising performance differences.

%%%

\paragraph{Multi-objective optimisation problems.}

So far, there are no systematic studies of automated algorithm selection for multi-objective continuous optimisation problems, as feature design is already extremely challenging and the suitability of existing benchmark sets is questionable. 
However, initial approaches regarding multi-objective features, landscape analysis and multimodality exist and offer promising perspectives for future research (see Section \ref{sec:perspectives}).

\section{Perspectives and Open Problems}
\label{sec:perspectives}

In the previous sections, we have given an overview of the state-of-the-art in algorithm selection for discrete and continuous problems. 
We have summarised the general approach of per-instance algorithm selection and discussed a number of related problems, including per-set algorithm selection, automated algorithm configuration, algorithm schedules and parallel algorithm portfolios.
An important aspect for any per-instance algorithm selection approach is the design of informative and cheaply computable features that are able to characterise and differentiate problem instances w.r.t.~a given set of algorithms. 
We have given an overview of feature sets for several prominent discrete problems, as well as features used in continuous black-box optimisation. 
Based on this, we have summarised prominent algorithm selection approaches for discrete and continuous problems. In the following, we will discuss perspectives and challenges for future research in the area of automated algorithm selection.

%%%

\paragraph{Performance measures.}

A crucial part of any empirical performance evaluation -- which provides the basis for constructing algorithm selectors -- is the underlying performance measure. 
While \textit{penalised average running time} \citep[notably, PAR10,][]{bischl2016_aslib} and \textit{expected running time} \citep[ERT,][]{hansen2009} are commonly used in this context,
some of its parameters can substantially affect performance measurements.
For example, in the case of PAR10, the penalty factor is set to 10 and an arithmetic mean is used to aggregate over multiple runs or problem instances. 
Both choices can, in principle, be varied (\eg{}, by replacing the arithmetic mean by different quantiles), with significant effects on the robustness of solvers selected based on them. 

Recently, \cite{kerschke2018} presented a structured approach on how to assess the sensitivity of an empirical performance evaluation w.r.t.~altered requirements regarding solver robustness across runs, focusing on PAR10 and ERT. 
As demonstrated within this study, by adjusting the parameters of the performance measures, users are able to adapt the ranking of a given set of algorithms, and the performance characteristics of the algorithm selectors constructed based on that set, according to their preferences -- trading off, for example, running time \vs{}~robustness.
In an alternative approach, \cite{vanrijn2017} utilise the advantages of two popular performance measures -- ERT and \textit{fixed cost error} \citep[FCE, see, \eg{}][]{baeck2013} -- by combining and standardising them within a joint performance measure. 
Moreover, a multi-objective perspective on performance measurement shows promise, \eg{}, by enabling direct investigations of the trade-off between the number of failed runs and the average running time of successful runs of a given solver \citep{bossek2018}. 
Concepts such as Pareto-optimality and related multi-objective quality indicators \citep{coello2007} could then be used in the context of constructing and assessing the performance of algorithm selectors and related meta-algorithmic techniques.

\paragraph{Evolving / generating problem instances.}

Naturally, an algorithm selector only generalises to problems which are similar enough to the instances contained in the benchmark set that was used to construct it. 
Usually, common benchmark sets are considered, which are deemed to be representative of a specific use context. However, one could argue that benchmark sets should be designed specifically w.r.t.~the given set of candidate solvers, such that they exhibit maximum diversity regarding the challenges posed by the instances for the specific solvers.
The idea of evolving instances utilising an evolutionary algorithm dates back to \cite{hemert2006} and \cite{smithmiles2010}, who constructed instances that are extremely hard or easy for specific TSP solvers; these works were followed by more sophisticated approaches of \cite{mersmann2012,mersmann2013} and \cite{nallaperuma2013}. \cite{bossek2016_evolving,bossek2016_understanding} further explored this idea by evolving TSP instances that maximise the performance difference between two given solvers, \ie{}, instances that are extremely hard to solve for one solver, but very easy for the other. 
This can, in principle, provide insights into links between performance differences and instance characteristics -- a topic that is highly relevant for automated algorithm selection. 

Recent studies build upon these concepts by explicitly focusing on the diversity of evolved instances \citep{gao2015,neumann2018}, paving the way for a systematic approach to construct most informative and relevant benchmarks specifically tailored to a given set of solvers. 
The next promising step in this direction will be the complementation of state-of-the-art benchmarks with those specifically designed instances in order to provide most informative benchmark sets tailored to given sets of solvers.
Such sets are likely to provide a basis for constructing per-instance algorithm selectors whose performance generalises better to problem instances that differ from those used during their construction.
Improvements, configuration and enhancements of the underlying evolutionary algorithm offer extremely promising research perspectives. 
Interestingly, the overall approach also provides a way for systematically detecting advantages and shortcomings of specific solvers, and thus produce benefits for the analysis and design of algorithms beyond algorithm selection and related approaches.

\paragraph{Online algorithm selection.}

The work covered in this survey is mainly related to (static) \textit{offline} algorithm selection, where algorithm selectors are constructed based on using a set of training instances \textit{prior} to applying them to new problem instances. 
Yet, according to \cite{armstrong2006}, \cite{gagliolo2010} and \cite{degroote2016}, in principle, it might be possible to obtain even better results using (dynamic) \textit{online} algorithm selection methods, which adapt an algorithm selector while it is being used to solve a series of problem instances.
Although this involves a certain overhead, it can enable better performance and increased robustness, as the selector can react better to changes in the stream of problem instances it is tasked to solve.
In order to more easily amortise the overhead involved in online algorithm selection, building on earlier work on probing features~\citep{hutter2014,kotthoff2015}, the development of cheap and informative monitoring features -- \ie{}, features that extract sufficient instance-specific information without significantly reducing solver or selector performance -- is likely to be of key importance.

Another approach for online algorithm selection and the selection of an appropriate algorithm for a given problem instance is provided by so-called \textit{hyper-heuristics}; these are algorithms for selecting or generating solvers for a given problem from a set of given heuristic components~\citep{burke2013}.
In many cases, they employ heuristic, rule-based mechanisms and make use of rather simple components, such as greedy construction algorithms for the given problem.

Life-long learning hyper-heuristics are applied in a setting where a series of problem instances is solved.
Based on the performance of previously chosen component solvers, a solver for a new problem instance is chosen that is deemed most likely to perform best.
Life-long learning hyper-heuristics have achieved good results for well-known decision problems, such as constraint satisfaction~\citep{ortiz2015} and bin packing~\citep{sim2015}.

\paragraph{Features for mixed (discrete + continuous) problems.}

Developing good features for a given problem is a challenging task that provides a crucial basis for effective algorithm selection techniques. 
As discussed previously, rich sets of features have been introduced for well-studied discrete and continuous problems, but many combinatorial problems of practical importance involve discrete and continuous decision variables.
Perhaps the best example for this is \textit{mixed integer programming (MIP)}, a problem of great importance in the context of a broad range of challenging real-world optimisation tasks.
\cite{hutter2014} introduced 95 features for MIP, including problem type and size, variable-constraint graph, linear constraint matrix, objective function and LP-based features.
For the \textit{travelling thief problem (TTP)}, which can be seen as a combination of the TSP and \textit{knapsack problem (KP)}, algorithm selection has been studied by~\cite{wagner2017}. 
They used 48 TSP and four KP features, plus three parameters of the TTP that connect the TSP and KP parts of the problem. Features for the KP that characterise the correlation of weights and profits have so far not been taken into account, although they seem to provide a further improvement in the characterisation of TTP instances.

\paragraph{Algorithm selection for multi-objective optimisation problems.}

While we have covered numerous studies on algorithm selection in this survey, none of them has dealt with multi-objective optimisation problems. 
From a practitioner's point of view, this is a significant limitation, as multiple competing objectives arise in many, perhaps most, real-world problems\footnote{Consider, for example, the trade-off between the performance improvement and the accompanying costs for finding such an improved solution.}.
However, for convenience, these problems are often handled as single-objective problems -- \eg{}, by focusing on the most important objective or by applying scalarisation functions.
Many prominent benchmarks for continuous multi-objective optimisation algorithms, such as bi-objective BBOB~\citep{tusar2016}, DTLZ~\citep{deb2005}, ED~\citep{emmerich2007}, MOP~\citep{vanveldhuizen1999}, UF~\citep{zhang2008}, WFG~\citep{huband2006} and ZDT~\citep{zitzler2000}, are entirely artificial, and it is unclear to which degree they resemble real-world problems.

In addition, there is a dearth of research on the characterisation of those optimisation problems -- in particular by means of automatically computable features.
Of course, one could compute variants of existing features for each of the objectives separately \citep[see, \eg{},][]{kerschke2016_flacco}, but this completely ignores interaction effects between the objectives, which in turn have a strong impact on the landscapes \citep[even for rather simple problems, see, \eg{},][]{kerschke2017_emo}. 
Hence, there is a significant need and opportunity for research on visualising multi-objective landscapes \citeg{fonseca1995, tusar2014, tusar2015, kerschke2017_emo}, as well as characterising them (numerically) -- along the lines of \cite{ulrich2010}, \cite{kerschke2016_ppsn,kerschke2018search} or \cite{grimme2018} -- as this will (a) improve the understanding of multi-objective problems and their specific properties, and (b) provide a basis for automated feature computation.
We expect the latter to be of key importance for the development of new algorithms and effective per-instance selectors in the area of multi-objective optimisation.

In order to construct a powerful and complementary portfolio of multi-objective optimisers (\textit{a priori}), as well as for analysing the strengths and weaknesses of the resulting algorithm selector (\textit{a posteriori}), visual approaches such as the \textit{empirical attainment function} (EAF) plots \citep{fonseca2005,fonseca2010,lopez2010} provide valuable feedback on location, spread and inter-point dependence structure of the considered optimisers' Pareto set approximations.

\paragraph{Algorithm selection on streaming data.}

The importance of automated algorithm selection and related approaches in the context of learning on streaming data should not be neglected.
Streaming data \citep{bifet2018} pose considerable challenges for the respective algorithms, as
(a) data points arrive as a constant stream,
(b) the size of the stream is large and potentially unbounded,
(c) the order of data points cannot be influenced,
(d) data points can typically only be evaluated once and are discarded afterwards, and
(e) the underlying distribution of the data points in the stream can change over time (non-stationarity or concept drift).

Few concepts for automated algorithm selection on streaming data exist so far, both for supervised \citeg{vanrijn2014, vanrijn2018} and unsupervised learning algorithms. In unsupervised learning, stream clustering is a very active research field. Although several stream clustering approaches exist \citep[see surveys of][]{amini2014, carnein2018, mansalis2018}, these have many parameters that affect their performance, yet clear guidelines on how to set and adjust them over time are lacking. Moreover, different kinds of algorithms are required in the so-called online phase (maintaining an informative aggregated data stream representation, in terms of microclusters) and offline phase (standard clustering algorithm, such as $k$-means applied to the microclusters), which leads to a huge space of parameter and algorithm combinations. 
An initial approach on configuring and benchmarking stream clustering approaches based on \textit{irace} \citep{lopez2016} has been presented by \cite{carnein2017}.
Building on this work, especially in light of remark (e) above, we expect that algorithm selection will play an important role for robust learning on streaming data -- especially when keeping the efficiency regarding real-time capability in mind. This will require informative feature sets, ensemble techniques, as well as a much wider range of suitable benchmarks. Of course, ideally a combination of algorithm selection and (online) configuration is desired and a very promising line of research.

\paragraph{Per-instance algorithm configuration.}

As previously explained, automatic algorithm configuration involves the determination of parameter settings of a given target algorithm to optimise its performance on a given set of problem instances. 
\emph{Per-instance algorithm configuration (PIAC)} is a variant of this problem in which parameter settings are determined for a given problem instance to be solved. 
It can be seen as a generalisation of per-instance algorithm selection, in which the set of algorithms that form the basis for selection comprises all (valid) configurations of a single, parameterised algorithm.
Analogous to per-instance algorithm selection, PIAC involves two phases: an offline phase, during which a per-instance configurator is trained, and an online phase, in which this configurator is run on given problem instances.
During the latter, a configuration is determined based on features of the problem instance to be solved.
Standard, per-set algorithm configuration, in contrast, is an offline process that results in a single configuration that is subsequently used to solve problem instances presumed to resemble those used during training.
PIAC is challenging, because the spaces of (valid) configurations to select from is typically very large 
\citeg{hutter2014_aclib}, and compared to the size of these configuration spaces, any training data used during the offline construction of the per-instance configurator is necessarily sparse. 
In particular, for typical configuration scenarios, the training data would necessarily cover only a very small number of configurations, which makes it challenging to learn a mapping from instance features to configurations.

We consider PIAC to be a largely open problem, with significant potential for future work. 
There is some evidence in the literature that it may have significant benefits compared to the more established per-set configuration techniques. 
Notably, \citet{kadioglu2010} proposed a PIAC approach based on a combination of clustering and a standard, per-set algorithm configurator and reported promising results on several set covering, mixed integer programming and propositional satisfiability algorithms.

\paragraph{Further challenges.}

Due to the steady stream of work in algorithm selection and related areas, it is important to keep track of promising developments.
Of course, domain-related research networks such as COSEAL\footnote{\url{https://www.coseal.net/}} might relieve this challenging task to some degree, yet they will not be able to keep up with all developments within this fast-growing and productive community.
Instead, comparisons against state-of-the-art methods, which are of special significance in this context, are facilitated by benchmarking platforms and libraries, such as \textsc{ASLib}~\citep{bischl2016_aslib}, \textsc{ACLib}~\citep{hutter2014_aclib} and \textsc{HPOlib}~\citep{eggensperger2013} for algorithm selection, configuration and hyperparameter optimisation, respectively.
At the same time, it is important (a)~to promote and establish as best practice the use of these libraries, especially in the context of newly proposed methods for algorithm selection and related problems, and (b)~to maintain and expand these libraries, in order to ensure their continued relevance, \eg{}, by integrating scenarios for multi-objective and additional real-world problems.

The latter not only applies to the previously mentioned libraries, but also to broader benchmark collections for the underlying specific problems.
For example, recent studies have analysed the `footprints' of different continuous optimisation algorithms on common benchmarks; while \cite{munoz2017} focused on BBOB (arguably the most prominent benchmark in continuous black-box optimisation), \cite{munoz2018} applied a similar analysis to machine learning problems from the UCI repository~\citep{dua2017} and OpenML~\citep{vanrijn2013}. 
Their visual analyses indicate that (a)~different ``comfort zones'' for the various algorithms in question exist across the respective instance spaces, in line with what might be expected based on a liberal interpretation of the NFL theorems by \cite{wolpert1997}, and (b)~the instances from common benchmarks' problems in continuous optimisation are not very diverse, but cover only relatively small areas of the overall problem instance space.

Another important direction for future work is the improvement of problem-specific features in general. 
Aside from the directions outlined previously (monitoring features as well as features for mixed and multi-objective problems), more informative and cheaper features are always desirable and likely to pave the way towards more effective applications of algorithm selection and related techniques.

An interesting open question regards the trade-off between the performance achieved by algorithm selection approaches, \eg{}, in relation to a hypothetical perfect selector (VBS), and their complexity, including the complexity of the feature sets they operate on.
There is recent evidence from an application of algorithm selection to solvers for quantified Boolean formulae (QBF) that suggests that sometimes, a small number of simple features is sufficient for achieving excellent performance \citep{hoos2018}.
However, it is presently unclear to which extent this situation arises in other application scenarios, and to which degree it is contingent on the use of highly sophisticated algorithm selection techniques.

Finally, an intriguing direction for future work is the development of algorithm selection techniques for automated algorithm configurators and selectors.
Intuitively, it is clear that different algorithm configuration scenarios would be handled most efficiently using rather different 
configuration procedures (depending, \eg{}, on the prevalence of numerical \vs{}~categorical parameters).
Likewise, it has been observed in the recent \textit{Open Algorithm Selection Competition} \citep{lindauer2018}
that different AS techniques work best on different AS scenarios -- suggesting that meta-algorithm selection (\ie{}, AS applied to AS strategies) might be useful for quickly identifying the selection strategy to be used in a particular application context.
In both cases, configurator selection and meta-selection, the limited amount of training data is likely to give rise to specific challenges, which may well require the development of new AS techniques.

%%%%%% Acknowledgements
\subsection*{Acknowledgements}

Pascal Kerschke, Heike Trautmann and Holger H.~Hoos acknowledge support from the \textit{European Research Center for Information Systems (ERCIS)}. The former two also acknowledge support from the DAAD PPP project No.~57314626. Frank Neumann acknowledges the support of the Australian Research Council through grant DP160102401. The authors gratefully acknowledge useful inputs from Lars Kotthoff and Jakob Bossek.

%%%%%% Bibliography
\bibliographystyle{apalike}
\bibliography{ecj_si}

\end{document}